\documentclass[a4paper,twoside]{article}

\usepackage{epsfig}
\usepackage{subcaption}
\usepackage{calc}
\usepackage{amssymb}
\usepackage{amstext}
\usepackage{amsmath}
\usepackage{amsthm}
\usepackage{multicol}
\usepackage{pslatex}
\usepackage{apalike}
\usepackage[bottom]{footmisc}

\usepackage{times}
\usepackage{epsfig}
\usepackage{graphicx}
\usepackage{amsmath}
\usepackage{amssymb}
\usepackage{tikz}
\usepackage{multirow}
\usepackage{booktabs}
\usepackage{hyperref}
\usepackage{cleveref}

\usepackage{pifont}

\usepackage{SCITEPRESS}     

\begin{document}

\title{Fake it, Mix it, Segment it:\\ Bridging the Domain Gap Between Lidar Sensors}

\author{\authorname{Frederik Hasecke\sup{1, 2}\orcidAuthor{0000-0002-6724-5649}, Pascal Colling\sup{2}\orcidAuthor{0000-0001-5599-1786} and Anton Kummert\sup{1}\orcidAuthor{0000-0002-0282-5087}}
\affiliation{\sup{1}Faculty of Electrical Engineering, University of Wuppertal, Germany}
\affiliation{\sup{2}Department of Artificial Intelligence and Machine Learning, Aptiv, Wuppertal, Germany}
\email{\{frederik.hasecke, kummert\}@uni-wuppertal.de, \{frederik.hasecke, pascal.colling\}@aptiv.com}
}


\abstract{Segmentation of lidar data is a task that provides rich, point-wise information about the environment of robots or autonomous vehicles. Currently best performing neural networks for lidar segmentation are fine-tuned to specific datasets. Switching the lidar sensor without retraining on a big set of annotated data from the new sensor creates a domain shift, which causes the network performance to drop drastically. In this work we propose a new method for lidar domain adaption, in which we use annotated panoptic lidar datasets and recreate the recorded scenes in the structure of a different lidar sensor. We narrow the domain gap to the target data by recreating panoptic data from one domain in another and mixing the generated data with parts of (pseudo) labeled target domain data. 
Our method improves the nuScenes \cite{caesar2020nuscenes} to SemanticKITTI \cite{behley2019semantickitti} unsupervised domain adaptation performance by $15.2$ mean Intersection over Union points (mIoU) and by $48.3$ mIoU in our semi-supervised approach.
We demonstrate a similar improvement for the SemanticKITTI to nuScenes domain adaptation by $21.8$ mIoU and $51.5$ mIoU, respectively. We compare our method with two state of the art approaches for semantic lidar segmentation domain adaptation with a significant improvement for unsupervised and semi-supervised domain adaptation. Furthermore we successfully apply our proposed method to two entirely unlabeled datasets of two state of the art lidar sensors \textit{Velodyne Alpha Prime} and \textit{InnovizTwo}, and train well performing semantic segmentation networks for both.}

\onecolumn \maketitle \normalsize \setcounter{footnote}{0} \vfill

\section{\uppercase{Introduction}}
\label{section:intro}
Lidar point cloud segmentation has grown immensely in importance in recent years. Unlike 3D bounding box annotations, segmentation cannot only provide information about other road users and specific static objects, but also convey precise information about the position of each individual data point as well as its relation to other points, the underlying geometry as well as the semantic meaning. In short, segmentation provides a complete picture of the real, underlying environment.
\begin{figure}[h!]
	\vspace{4pt}
	\centering
	\includegraphics[width=0.45\textwidth]{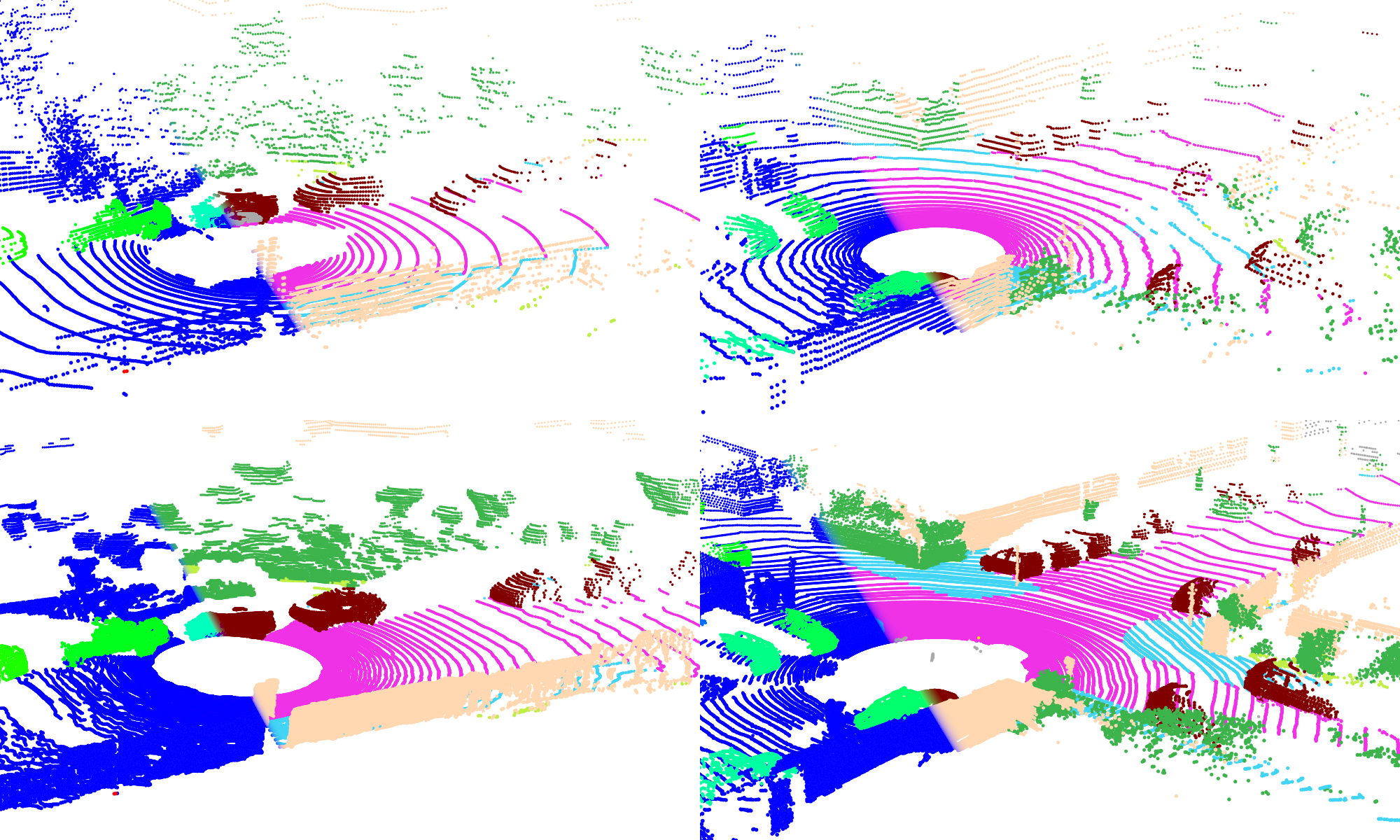}
	\caption{\textbf{Panoptic Lidar Point Clouds and Their Respective Twins in a Different Lidar Sensor Domain.}\\ We modified the lidar structure and the existing classes of both datasets so that they exist in both domains: Real nuScenes (top left) as fake SemanticKITTI (bottom left), real SemanticKITTI (bottom right) as fake nuScenes (top right). Best viewed in color on a digital device, color map explained in \textsc{Figure} \ref{fig:class_mapping}}
	\label{fig:eyecatcher}
\end{figure}
This richness of data allows decision algorithms of autonomous vehicles and robots to have a complete understanding of the environment, and thus make decisions that are not based on a reduced dataset of boxes, for instance. However from the advantages of this immense amount of data through segmentation stems the main challenge: 
a segmentation model with high accuracy requires annotated data, i.e., that data has to be annotated by humans.
Unlike box annotations, segmentation annotations are a tedious and costly effort. Not without reason, there are many more public datasets with 3D box annotations \cite{geiger2012cvpr}\cite{caesar2020nuscenes}\cite{sun2020scalability}\cite{lyft2019}\cite{innovizeccv} than those with full segmentations of lidar point clouds \cite{behley2019semantickitti}\cite{fong2021panoptic}\cite{xiao2021pandaset}. 
Another problem is the specificity of the different datasets. Lidar data is much more difficult to combine than for example image data. The different designs of lidar sensors and the different mounting positions make it infeasible to generalize between existing annotated and new unlabeled datasets. Lidar data can rarely if ever be reused for a different application. Current state of the art domain adaptation methods for lidar segmentation use alignment of geometric  and feature statistics at the data level \cite{alonso2020domain}\cite{rochan2022unsupervised}, and use network specific adaptations at the model level to reduce the domain shift between datasets \cite{bevsic2022unsupervised}\cite{corral2021lidar}. 

Our approach, on the other hand, works exclusively at the data level to align different lidar domains, and we deliberately choose not to align at the model level for a more general approach. Our method bridges the domain gap by applying sensor structure aware domain adaptation modules that mix source and target data by using self- and semi-supervised data fusion methods. 
For this, we combine the point clouds of a panoptic source dataset into a static mesh world and ray-trace the mesh with a virtual target lidar twin, to recreate the data in the structure of the target sensor, as shown in \textsc{Figure} \ref{fig:eyecatcher}. Furthermore, we extend this unsupervised domain adaptation with additional semi-supervised and self-supervised approaches to mitigate the domain shift between datasets to such an extent that we can train competitive lidar segmentation networks.



\section{\uppercase{Related Works}}
\label{section:related}

\subsection{Lidar Segmentation}
In recent years, the state of the art in lidar segmentation has changed dramatically. Early lidar segmentations mostly focused on foreground classification and clustering of individual objects  \cite{moosmann2009segmentation}\cite{Bogoslavskyi2016}. Early semantic segmentation networks extracted point-wise classes from cuboid label datasets to perform simple semantic segmentation of foreground classes \cite{wu2018squeezeseg}. With the release of the semantic label extension of the original KITTI data \cite{geiger2012cvpr}, the SemanticKITTI dataset \cite{behley2019semantickitti}, which covers a variety of foreground and background classes, a new wave of semantic segmentation algorithms for lidar data emerged. The segmentation networks evolved from range image projection networks \cite{milioto2019rangenet++}\cite{cortinhal2020salsanext}, to point-based segmentation \cite{thomas2019kpconv}, to voxel-based \cite{tang2020searching}\cite{zhou2020cylinder3d}, networks, to combining several working principles \cite{hou2022point}\cite{xu2021rpvnet}, and even multi-modalities by combing the lidar data with camera data \cite{yan20222dpass}. 
In the meantime, the authors of nuScenes \cite{caesar2020nuscenes} have also extended their dataset with point-based semantic and panoptic segmentation \cite{fong2021panoptic}, and built a network performance competition for this data. These two datasets represent the most used datasets for lidar segmentation to date. Both are used in this work.

\subsection{Simulation}
\label{section:related:simulation}
One approach of lidar domain adaptation is the \textit{'simulation to real'} adaptation \cite{Dosovitskiy17}. The goal is a complete simulation of the sensor data in a computer program in order to create a large pool of annotated training data for a target sensor. Here, the physical principles of the lidar sensor are re-implemented in a virtual twin and moved through a virtual world. By inherently knowing the position of the virtual sensor, as well as the entire virtual environment, the annotation labels of the visible objects can be attached to each generated data point. 
The main problem that arises from simulated datasets is the domain shift to real data. Despite the very similar recording method of the virtual sensor twin to the real sensor, many simulators suffer from a too perfect mapping of the environment, as well as a too clean environment. The authors of \cite{xiao2022transfer}\cite{zhao2021epointda} have proposed data-level methods to adjust the appearance and sparsity of simulated point clouds to be more similar to real recordings. In \cite{saltori2022cosmix} the domain shift is addressed by adding parts of pseudo labeled real data to simulated data. 
Unfortunately, another downside of simulated environments is the simulation itself. It can only create scenarios as diverse as can be represented by the underlying simulation environment. 

\subsection{Domain Adaption}
\label{section:related:domain_adaption}
Following the publication of the aforementioned segmentation datasets, several approaches for \textit{'real to real'} data lidar domain adaptation have appeared. The authors of \cite{alonso2020domain} proposed a straight forward sensor to sensor domain adaptation by translating the source point cloud and removing lidar channels from the higher resolution sensor. In \cite{langer2020domain} and \cite{bevsic2022unsupervised}, the authors sum the point clouds over shorter and longer periods of time, to create a larger point cloud. Both works use mesh methods to fill in the gaps between the lidar points. While the authors of \cite{langer2020domain} achieve better performance based on point clouds without a mesh, the work in \cite{bevsic2022unsupervised} uses a combination of mesh objects for sparsely populated instances and point clouds for densely populated objects. The authors of \cite{yi2021complete} similarly reformulate the domain adaptation as a surface completion task by applying a Poisson surface reconstruction algorithm on a point cloud and ray-trace the surface with a virtual lidar. In \cite{jiang2021lidarnet} a pre-processing model is proposed, that uses label in-painting to bridge the sparse point labels in a range image projection. 
Other works use Generative Adversarial Networks \cite{corral2021lidar} or mask the range images \cite{rochan2022unsupervised} to make the range image projections of one dataset look like another, and use range image based networks for lidar semantic segmentation \cite{cortinhal2020salsanext}.

Our approach differentiates from the aforementioned works, as we combine the unsupervised recreation of real sensor data in a different lidar domain together with novel fusion methods of generated and real target data. Our method bridges the gap between the domains and thus enables competitive segmentation networks with minimal annotations for the target lidar sensor. Our approach also works without any target lidar segmentation labels due to our novel combination of self-supervised pseudo labels with the generated point clouds.

\section{\uppercase{Method}}
\label{section:method}
\begin{figure*}[tbh!]
	\vspace{4pt}
	\centering
	\includegraphics[width=0.9\textwidth]{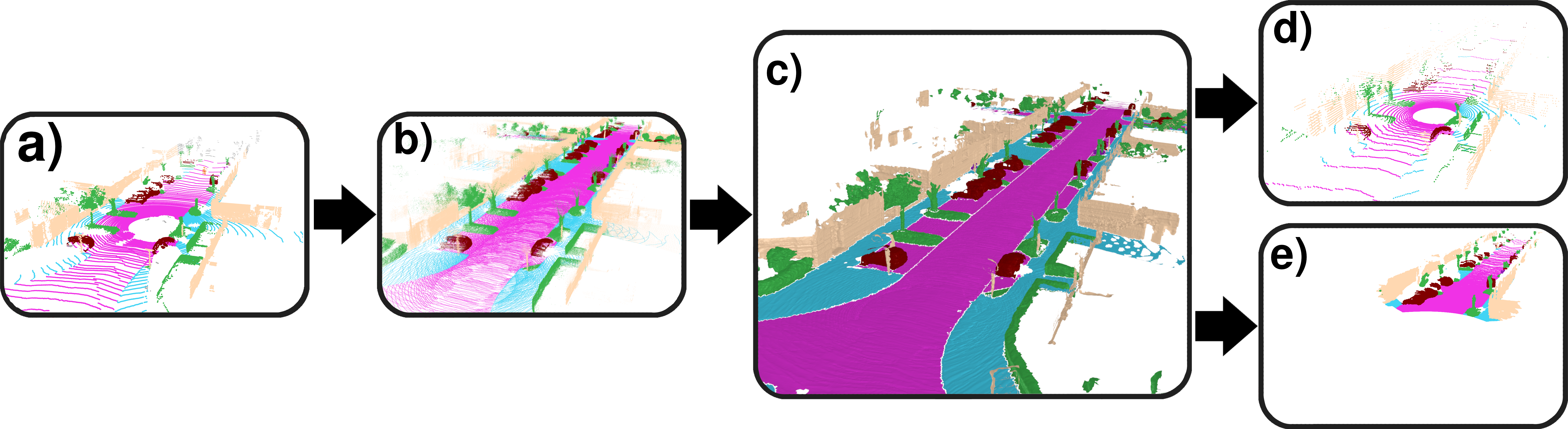}
	\caption{\textbf{Restructuring of a Single Dataset in the Form of Several Different Sensors.} We use the SemanticKITTI dataset (a), sum up all point clouds (b) create a mesh world (c) and retrace the lidar structure of the \textit{VLP-32C} (d) used in the nuScenes dataset as well as the \textit{InnovizTwo} lidar sensor (e). Best viewed in color on a digital device.}
	\label{fig:mesh_tracing}
\end{figure*}
We propose a data-centric approach to panoptic lidar domain adaptation that conserves the semantic and instance labels of the source dataset.  We recreate the entire scene in the shape, range and structure of any other lidar sensor to accommodate all types of segmentation network and to be able to train on the resulting data. 
We use the entire sequences of the source dataset to create the static underlying environment in the structure of the target sensor. Then we fill these static scenes with dynamic objects and reduce the domain shift between the generated data and real data of the target sensor. For this we use small pools of annotated data or pseudo labeled data of previous inference iterations of trained networks.

\subsection{Non-Causal Data Collection}
\label{section:method:scene2frame}
In order to bridge the gaps between adjacent measurements of a single lidar scan, we summarize all points of a sequential scene of our source dataset. This provides us with a denser representation of the underlying real world scene, that the source dataset captured and annotated. While both datasets SemanticKITTI \cite{behley2019semantickitti}\cite{geiger2012cvpr} and nuScenes \cite{caesar2020nuscenes} provide the ego-motion ground truth for the training and validation data, we can extend our approach to other datasets by utilizing various SLAM algorithms. In order to prevent dynamic instances such as driving cars and moving pedestrians to smear across the static point map, we remove all dynamic instances from all point scenes. The attribute of dynamic objects is given for the two used source datasets. The resulting scene point clouds of the source data appears much denser, but the points are still zero dimensional point probes (compare \textsc{Figure} \ref{fig:mesh_tracing} b). In order to sub-select or ray-trace the scene point cloud in the structure of the target lidar sensors one has to use e.g. closest-point sampling \cite{langer2020domain}. 
The resulting point cloud exhibits a visual structure close to the target lidar sensor, but can not provide information between two points that are too far apart, sampling methods also tend to introduce unrealistic representations such as visible points behind walls or other objects due to missing direct occlusions resulting from the zero-dimensional nature of the points \cite{langer2020domain}. We therefore decided to fill these gaps with a mesh representation derived from the scene point cloud.

\subsection{Lidar Mesh Creation}
\label{section:method:mesh}
Recreating a surface model from point clouds has been studied for close to a century \cite{delaunay1934sphere}. Well known methods include the alpha shapes algorithm \cite{edelsbrunner1983shape}, the truncated signed distance function \cite{curless1996volumetric} and the Poisson surface reconstruction algorithm \cite{kazhdan2006poisson}, to name a few. We choose the Open3D \cite{Zhou2018} implementation of the latter to recreate the scene point cloud as a scene mesh object. We decided for this due to the ease of use and the triviality of the required normal vector estimation \cite{behley2018efficient}.
In a second step, we used a \textit{k}-nearest neighbors sampling \cite{fix1989discriminatory} to assign each mesh vertex a class label, an instance label as well as a mean intensity reflection value. Here we took the 10 nearest neighbors of each vertex in the original scene point cloud, and assigned the most frequent value for the class label as well as the instance label. The intensity value reflects the mean value of the 10 nearest original points, with an inverse distance weighting.

\subsection{Virtual Lidar Sampling}
\label{section:method:sampling}
To recreate a new point cloud from the mesh object in the structure of the target lidar sensor, we choose the raycasting method. It is advantageous that the rotating lidar sensors of most automotive datasets have a relatively uniform vertical distribution of laser depth measurement modules. By rotating around a common axis, the individual measurement channels also have a uniform horizontal resolution. This peculiarity allows us to use a more efficient way of raycasting, where we do not have to trace the entire mesh model in three dimensions, but we can rely on proven methods of 3D computer graphics rendering. 

We project the mesh environment from Cartesian to spherical coordinates to capture a depth image from the perspective of the lidar sensor by using a virtual orthographic camera. Other adjustments such as the sensor location and rotation are applied to the orthographic camera. We choose a render resolution that is three times the lidar resolution, which we then subsample to arrive at the resolution of the target sensor. We decided on this to mitigate excessive scaling of distant objects by discretizing a fixed grid image. The mesh rendering allows us to mirror the concept of a single dimensional measurement beam relatively accurately with a two dimensional depth image. Since each pixel position has a well-defined depth, azimuth and an elevation angle, we can reformulate these three values into a Cartesian coordinate system to obtain a pseudo lidar point cloud in the structure of the target sensor. In addition, we take semantic, instance and reflection values directly from the mesh model to assign them to the newly created points. With this simple and efficient approach, we can use a single mesh world to recreate the structure of any number of different lidar sensors. In \textsc{Figure} \ref{fig:mesh_tracing}, we show how two completely different sensor point cloud are recreated from a lidar mesh world that we created out of the SemanticKITTI dataset.

\subsection{Instance Injections}
\label{section:method:semi:injections}
The method described above creates very accurate representations of the source data in the structure of the target data. However, a serious problem arises here. The generated scenes represent only the static components of the source data. Driving cars, bicyclists and pedestrians are filtered out by our non-causal collection method, so that the resulting scenes all represent empty streets without any traffic.
\begin{figure}[tb]
	\vspace{4pt}
	\centering
	\includegraphics[width=0.45\textwidth]{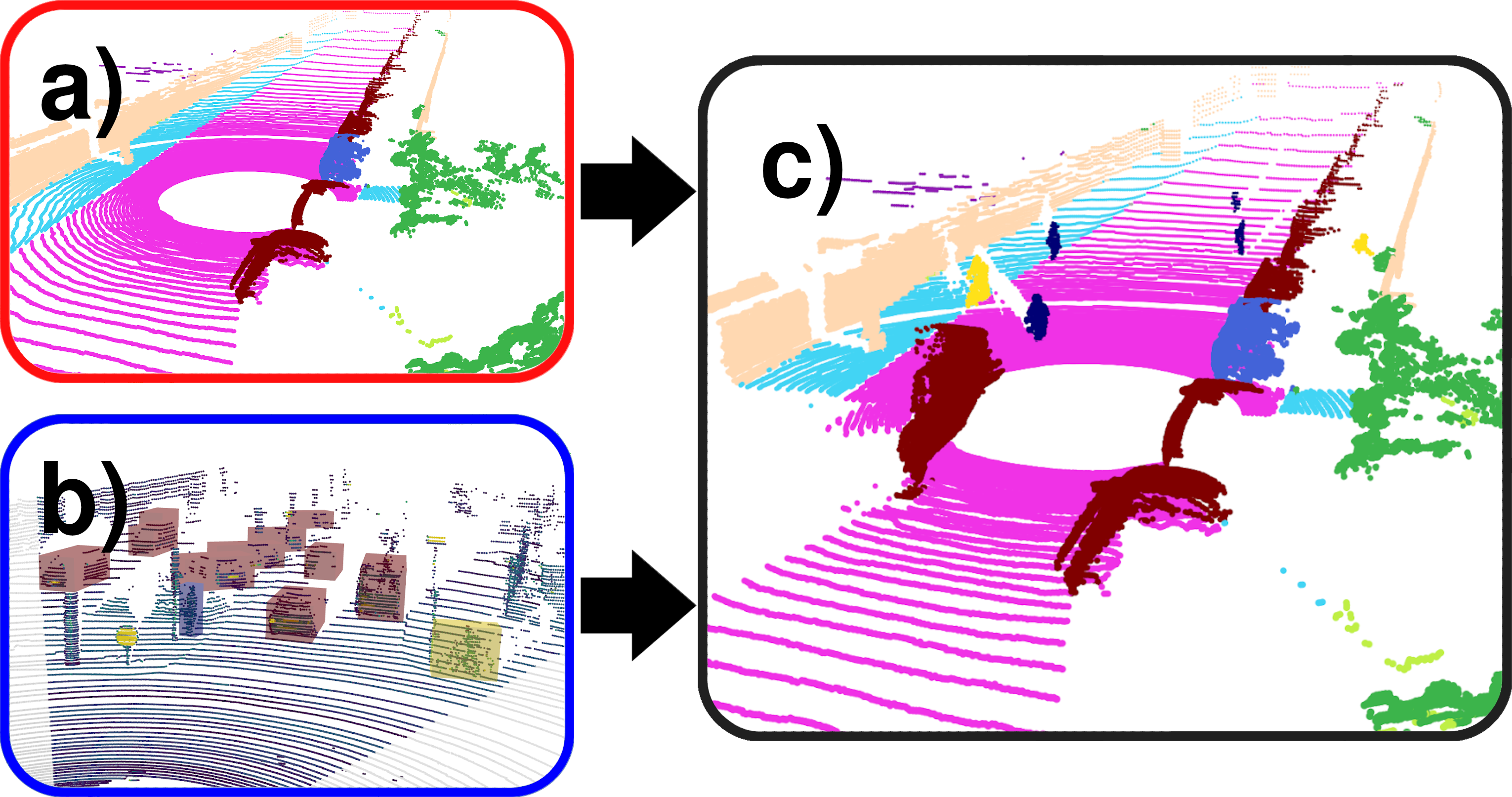}
	\caption{\textbf{Target Domain Data Injection into Generated Lidar Point Clouds.} We combine our generated static scene (a) with sampled target sensor (pseudo) ground truth data (b), that we extract from cuboid labels or alternatively bounding box predictions. We inject the instances into the generated scenes to create dynamic lidar data (c) consisting of parts of the source and the target domain.}
	\label{fig:injection}
\end{figure}
To fix this problem, we created an additional step, to bring these empty scenes back to life. In our simple semi-supervised approach, that is, without using any ground truth data for the target dataset, we apply general purpose object detectors to the unlabeled target lidar data. The points within a box prediction - together with the semantic and instance label of the box - are then cut out of the scene and inserted into the empty, recreated segmentation scenes as dynamic objects as shown in \textsc{Figure} \ref{fig:injection}. Alternatively we can use the same method with ground truth cuboid labels if these are available for the target data.
Inserting bounding box labels as segmented point-wise instances has a triple benefit. First, dynamic objects are inserted back into the static scene. Second, we disproportionately adjust the distribution of underrepresented classes to force our segmentation networks to see these classes more often and adjust the weights to them accordingly. And third, by mixing generated scene point clouds and real instance point clouds, we bring the two separate distributions of the real and generated domains closer together to narrow the gap between the two domains. For the semi-supervised approaches in \textsc{Section} \ref{section:experiments:n2k} and \ref{section:experiments:k2n} we use a subset of the provided bounding box labels from the KITTI \cite{geiger2012cvpr} and nuScenes \cite{caesar2020nuscenes} dataset respectively for the injection of instances.

\subsection{Mixing Domains}
\label{section:method:semi:mixture}
Recently multiple lidar augmentation methods have been published, that go beyond the injection of single objects into a scene, to a complete mixture between two lidar point clouds recorded at different positions and different times. Mix3D \cite{nekrasov2021mix3d} proposes the straight forward concatenation of two point clouds, in order to break up the context of certain classes and objects. 
A similar approach was proposed by the authors of \cite{hasecke2022can}, but they kept only parts of each point cloud according to their distance to the lidar sensor, thus creating a mixed point cloud while keeping the structure of the lidar sensor intact. 
\begin{figure*}[tb]
	\vspace{4pt}
	\centering
	\includegraphics[width=0.9\textwidth]{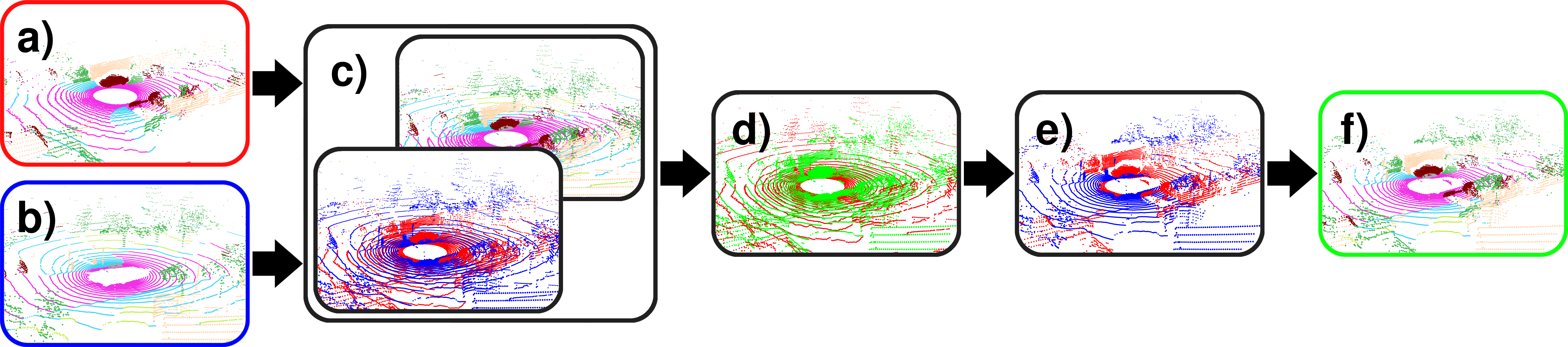}
	\caption{\textbf{Point-wise Domain Fusion by Range.} We select a generated lidar scene (a) and a (pseudo) ground truth lidar frame of the target sensor (b). We move both frames into the same origin (c) and apply the point-wise range competition (d) in the range image domain: the green points are closer to the lidar sensor. We end up with a new point cloud (e) exhibiting parts of the real target data (blue) as well as the labeled generated data (red). The final result (f) is a structurally intact point cloud consisting of both generated and real data. Graph adapted from \cite{hasecke2022can}, best viewed in color.}
	\label{fig:fusion}
\end{figure*}
We base our domain mixing approach on the latter. Instead of fusing point clouds of the same dataset to decrease the effect of overfitting, we combine our synthetic generated scenes, created with the method from \textsc{Section} \ref{section:method:sampling} with a subset of target lidar data, as can be seen in \textsc{Figure} \ref{fig:fusion}: The basic principle is best explained by the semi-supervised approach, but can also be applied to pseudo labeled lidar frames. We mix a very small subset of real, annotated data of the target dataset as a separate data source to our generated scenes. Both data pools, the real data as well as the generated data, exhibit the same lidar characteristics. 
The blending of these two sources not only increases the diversity of the overall dataset, but also interpolates the two domains within a single point cloud, thus reducing the domain shift between the two. The authors of \cite{saltori2022cosmix} noticed a similar effect; merging patches of different domain sources pull them closer together in the total distribution. Our method increases this pull effect due to the structure aware fusion of the different point clouds.

\subsection{Pseudo Labels}
\label{section:method:pseudo}
The previously mentioned effect of pulling domains together in order to lessen the shift between them can be applied in both, a semi-supervised but also in an unsupervised fashion. For this, we use a network trained on the domain adapted data in order to create pseudo labels for unlabeled data of the target domain. In a second step, we apply the same methods described above with the pseudo labeled data instead of a small annotated data pool. In order to reduce the influence of wrong labels in the final network, we remove all points with a probability of less then 85\%. The advantage of our reformulated fusion methods of \cite{hasecke2022can} to other pseudo label approaches is, that we do not produce empty point clouds when we remove these uncertain regions, but populate the empty positions with the complete scene point clouds of our generated samples. 

\section{\uppercase{Experiments}}
\label{section:experiments}
\begin{figure}[tb]
	\vspace{4pt}
	\centering
	\includegraphics[width=0.4\textwidth]{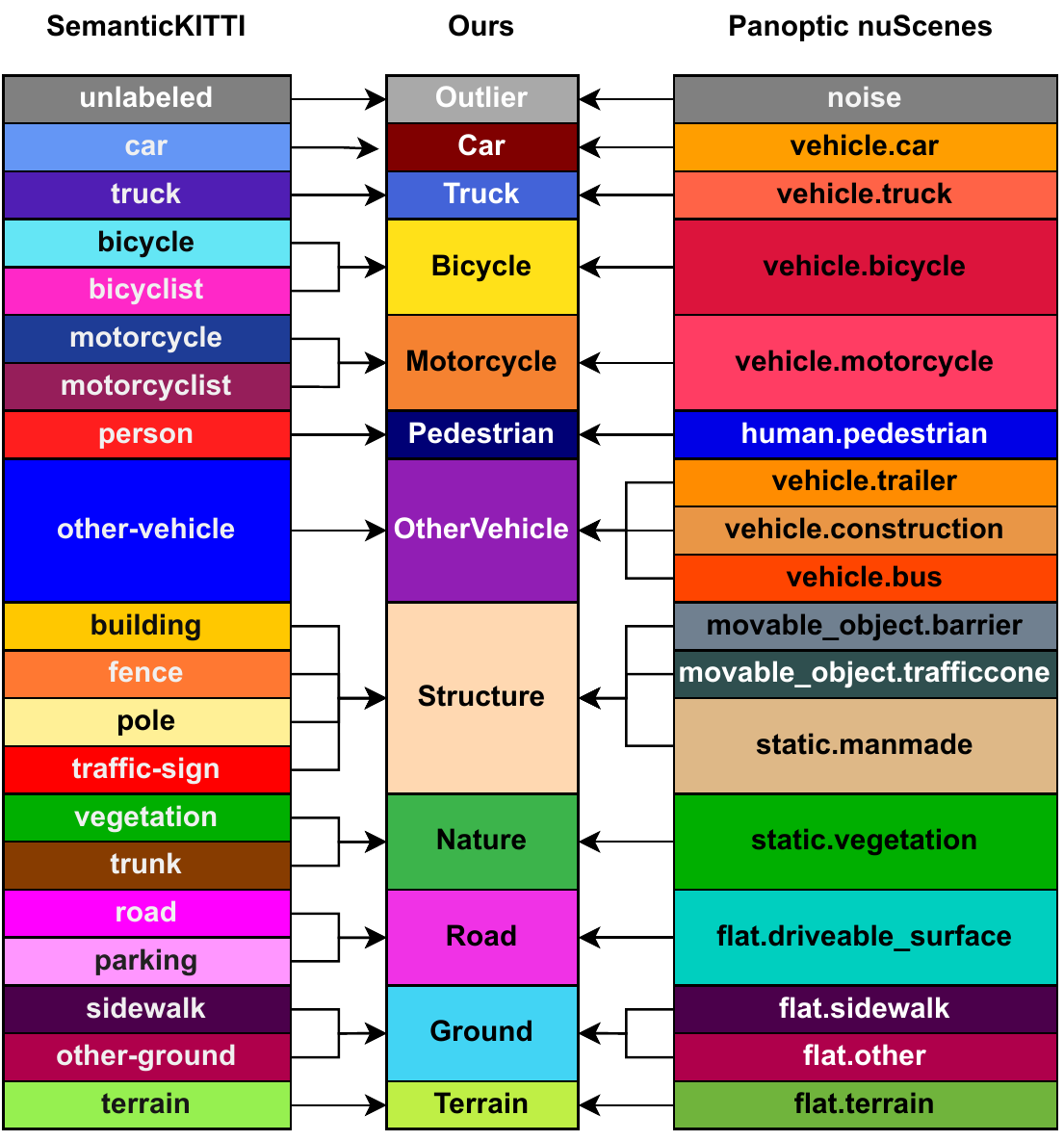}
	\caption{\textbf{Joint Class Mapping of the Datasets.} We remap the classes of both datasets used in this work to match the different classes in joint categories, that are present in both datasets for a uniform class label set.}
	\label{fig:class_mapping}
\end{figure}
In order to show the effectiveness of our lidar domain adaptation method we used two open source datasets: SemanticKITTI \cite{behley2019semantickitti}, which is a panoptic label extension to the KITTI \cite{geiger2012cvpr} odometry dataset and panoptic nuScenes \cite{fong2021panoptic} which provides panoptic segmentation labels to the nuScenes \cite{caesar2020nuscenes} dataset. The datasets use two different lidar sensors, mounted on different vehicles, at different heights and record data on different continents. In conclusion, the domain gap between the two is as large as possible while keeping the application of automotive lidar segmentation the same. To apply our domain adaptation method and enable a comparison of the performance between the two datasets, we first need to remap the classes present in both datasets. In total SemanticKITTI contains $28$ classes, including those which differentiate between non-moving and moving objects.
The authors recommend to remap these classes to a subset of $20$. NuScenes consists of $31$ classes which are in turn mapped to a subset of $17$. 

In order to compare the performance of segmentation networks between the two datasets we decided to remap both to the joint class set shown in \textsc{Figure} \ref{fig:class_mapping}. The authors of SemanticKITTI offer a suggestion for coarser category labels, that remap the original $28$ classes to the $7$ categories \textit{'ground', 'structure', 'vehicle', 'nature', 'human', 'object' and 'outlier'}, but we tried to keep as many classes as possible between the two datasets with our joint label mapping. Unfortunately, the use of different class combinations prevent direct comparison with most other methods \cite{langer2020domain}\cite{bevsic2022unsupervised} for Lidar Domain Adaptation for segmentation, therefore we only compare our method to \cite{corral2021lidar} and \cite{rochan2022unsupervised}.

\subsection{NuScenes to SemanticKITTI}
\label{section:experiments:n2k}

\begin{table*}[!t]
	\caption{\textbf{NuScenes to SemanticKITTI Ablation Study of our Domain Adaption Method Using the Cylinder3D Network \cite{zhou2020cylinder3d}}
	The classes are joined from the source and target dataset according to \textsc{Figure} \ref{fig:class_mapping}. 
	"GT Frames" denote the addition of a small subset of $100$ annotated target frames ($0.5 \%$ of the training data), while "GT Inst." is the addition of cuboid detections as point-wise labels. All Cylinder3D networks have been trained from scratch with the same parameters to ensure a fair evaluation. We compare our method to the unsupervised domain adaptation method of \cite{rochan2022unsupervised} and the semi-supervised domain adaptation of \cite{corral2021lidar} which uses $100$ annotated target frames for the training and list their reported IoU. \textbf{\textcolor{red}{Best results are shown in bold red}},  \textit{\textcolor{blue}{second best in italic blue text}}.}
	\resizebox{\textwidth}{!}{
		\begin{tabular}{l|cccc|c|ccccccccccc}
			\toprule
			& \rotatebox[origin=c]{75}{Gen. Frames} &  \rotatebox[origin=c]{75}{Pseudo Labels} & \rotatebox[origin=c]{75}{GT Inst.} & \rotatebox[origin=c]{75}{GT Frames} & \rotatebox[origin=c]{75}{mIoU} & \rotatebox[origin=c]{75}{Car} & \rotatebox[origin=c]{75}{Truck} & \rotatebox[origin=c]{75}{Bicycle} & \rotatebox[origin=c]{75}{Motorcycle} & \rotatebox[origin=c]{75}{Pedestrian} & \rotatebox[origin=c]{75}{OtherVehicle} & \rotatebox[origin=c]{75}{Structure} & \rotatebox[origin=c]{75}{Nature} & \rotatebox[origin=c]{75}{Road} & \rotatebox[origin=c]{75}{Ground} & \rotatebox[origin=c]{75}{Terrain} \\
			\midrule
			\midrule
			
			\multirow{3}{*}{Unsup. (Ours)}  &  & &  & & 19.1 & 64.5 & 0.9 & 0.0 & 5.0 & 0.0 & 1.0 & 38.3 & 11.0 & 50.6 & 4.8 & 33.7\\

			
			&\checkmark&  &  &  & 30.7 & 86.1 & 6.8 & 5.8 & 16.0 & 1.2 & 3.4 & 44.6 & 29.9 & 64.2 & 32.9 & 47.1 \\
			&\checkmark& \checkmark &  &  & 34.3 & 88.8 & 3.0 & 1.0 & 16.9 & 0.3 & 1.0 & 49.3 & 42.5 & 74.0 & 51.2 & 49.3 \\
			\midrule
			\multirow{3}{*}{Semi-Sup. (Ours)}	&\checkmark&  & \checkmark &  &   31.9 & 78.6 & 1.98 & 6.9 & 7.6 & 10.9 & 1.8 & 51.8 & 42.62 & 66.9 & 38.58 & 43.2 \\
			&\checkmark&  & \checkmark& \checkmark & 63.1 & 93.1 & 31.1 & 50.1 & 43.3 & 65.4 & 13.5 & 86.8 & 84.9 & 87.0 & 73.1 & 65.8\\
			
			&\checkmark& \checkmark & \checkmark& \checkmark & \textit{\textcolor{blue}{67.4}} & \textit{\textcolor{blue}{94.0}} & \textit{\textcolor{blue}{50.8}} & \textit{\textcolor{blue}{58.2}} & \textit{\textcolor{blue}{51.9}} & \textbf{\textcolor{red}{71.6}} & 13.9 & \textit{\textcolor{blue}{88.3}} & \textit{\textcolor{blue}{85.8}} & 88.2 & \textit{\textcolor{blue}{75.3}} & 67.0\\
			\midrule
			
			\cite{rochan2022unsupervised} & &  &  & & 23.5 & 49.6 & 1.8 & 4.6 & 6.3 & 12.5 & 2.0 & 65.7 & 57.9 & 82.2 & 29.6 & 34.0 \\
			
			\cite{corral2021lidar} & &  &  & \checkmark & 46.2 & 87.3 & 27.6 & 29.2 & 26.9 & 34.6 & \textit{\textcolor{blue}{24.4}} & 61.7 & 46.4 & 70.3 & 52.3 & 47.4 \\
			
			\midrule
			
			\multirow{2}{*}{Supervised} &  \multicolumn{3}{c}{100 Frames} & \checkmark & 49.0 & 91.2 & 1.6 & 8.1 & 2.6 & 30.1 & 6.0 & 83.3 & 85.3 & \textit{\textcolor{blue}{88.3}} & 73.3 & \textbf{\textcolor{red}{69.6}}\\
			
			& \multicolumn{3}{c}{Full Target Dataset $\dagger$} & & \textbf{\textcolor{red}{75.8}} & \textbf{\textcolor{red}{96.5}} & \textbf{\textcolor{red}{84.7}} & \textbf{\textcolor{red}{62.3}} & \textbf{\textcolor{red}{53.7}} & \textit{\textcolor{blue}{70.2}} & \textbf{\textcolor{red}{53.2}} & \textbf{\textcolor{red}{89.5}} & \textbf{\textcolor{red}{86.0}} & \textbf{\textcolor{red}{91.0}} & \textbf{\textcolor{red}{79.2}} & \textit{\textcolor{blue}{67.3}} \\
						

			\bottomrule
		\end{tabular}
	}
	\begin{minipage}{0.9\textwidth}\footnotesize
		\vspace{5pt}
		$\dagger$ The target baseline mIoU is higher than reported by the original authors, as we are using the reduced joint class set as shown in \textsc{Figure} \ref{fig:class_mapping} and therefore eliminate some of the bad performing classes from the evaluation.
	\end{minipage}
	\centering
	\label{tab:ablation_n2k_semseg_ablation}
\end{table*}
The nuScenes dataset provides panoptic lidar labels, instance-wise attributes for dynamic objects and an ego-motion ground truth. The presence of this data allows us to clean the lidar point clouds from the dynamic objects and to combine all the point clouds of a sequence by their ego-motion.  
The dataset itself is divided into several sub-sequences, each with a length of $20$ frames and an acquisition speed of $2$\,Hz, covering a span of $10$ seconds. Our goal is to recreate new panoptic segmentation lidar data from this data in the structure of \textit{Velodyne HDL-64E} lidar sensor data. For this we sum up all point clouds for each sequence and create a 3D mesh world using Poisson surface reconstruction as described in \textsc{Section} \ref{section:method:mesh}.
Using the spherical projection of this mesh, we can represent the recording structure of the target sensor by a simple orthographic camera. For this we define a minimum and maximum vertical angle, as well as a horizontal image resolution. Thus we recreate the static scenes in the lidar structure of the KITTI dataset. Please note, that while the created data does provides panoptic labels, we only use the semantic labels for our experiments for semantic segmentation.
We conducted an ablation study to evaluate the impact of all modules of our method: We started by replacing the original nuScenes data ($19.1$ mIoU) with the recreated lidar frames as described, which boosts the performance to $30.7$ mIoU. Then we used the trained network in order to create pseudo labels for unlabeled data of the target sensor, that we mix to our generated frames resulting in a total mIoU of $34.3$.
In further steps we increase the amount of semi-supervision by first adding the object detection cuboid labels of the original 3D object detection dataset \cite{geiger2012cvpr} as described in \textsc{Section} \ref{section:method:semi:injections} without the use of pseudo labels ($31.9$ mIoU). Next, for the semi-supervised domain mixture we sample $100$ frames from the target domain, in order to fuse the domains (see \textsc{Section} \ref{section:method:semi:mixture}). These $100$ frames are less than $0.5 \%$ of the original dataset and the fusion results in a huge performance gain to a mIoU of $63.1$. The final version of our semi-supervised method consists of a network trained on the recreated data with all the parts mentioned above together with the pseudo labels derived from the previous network inferred on the unlabeled target lidar data. We add these pseudo labeled point clouds as additional fusion point clouds as explained in \textsc{Section} \ref{section:method:pseudo}, which boosts our final network performance to a grand total of $67.4$ mIoU, which is $89 \%$ of the segmentation quality of the same network trained on the full target dataset ($75.8$ mIoU).
For a fair comparison, we train the same network only on the $100$ sampled frames of the target dataset used in our semi-supervised approach resulting in a mIoU of $49.0$. As can be seen in \textsc{Table} \ref{tab:ablation_n2k_semseg_ablation} our domain adaptation method, our domain injection and especially our domain fusion each have a noticeable impact on the final segmentation quality. 
Furthermore, we compare our method with two state of the art lidar domain adaptation methods for semantic segmentation. The first approach \cite{rochan2022unsupervised} is a unsupervised method that does not perform domain adaptation directly on the lidar points, but in the range image domain with a reported mIoU of $23.5$. We cannot rule out that the used network is partly responsible for the better performance of our method. However, this difference highlights the advantage of our general domain adaptation approach, since it is not limited to networks for range images. The second work we compare to our method is \cite{corral2021lidar}. The authors of this paper also use a semi-supervised approach by using parts of the annotated target dataset in addition to their domain adaptation, in which they report a mIoU of $46.2$ with the use of $100$ annotated frames of the target dataset. They report further approaches using up to $500$ frames of the target dataset, yet only reach a performance of $53.6$ mIoU. Our performance of $67.4$ mIoU using only $100$ ground truth frames shows the great advantage of our domain fusion and injection methods for narrowing the domain gap between the datasets.

%
%
%

\subsection{SemanticKITTI to NuScenes}
\label{section:experiments:k2n}
\begin{table*}[th!]
	\caption{\textbf{SemanticKITTI to NuScenes Domain Adaption Methods Using the Cylinder3D Network \cite{zhou2020cylinder3d}}\\ All Cylinder3D networks have been trained from scratch with the same parameters to ensure a fair validation, but we list the reported IoU of the cited previous work. 
	\textbf{\textcolor{red}{Best results are shown in bold red}},  \textit{\textcolor{blue}{second best in italic blue text}}.}
	\resizebox{\textwidth}{!}{
		\begin{tabular}{l|c|ccccccccccc}
			\toprule
			Method & mIoU & Car & Truck & Bicycle & Motorcycle & Pedestrian & OtherVehicle & Structure & Nature & Road & Ground & Terrain\\
			
			\midrule
			\midrule
			No Domain Adaption & 7.4 & 3.7 & 0.3 & 0.0 & 0.1 & 0.1 & 0.5 & 18.2 & 0.1 & 11.3 & 1.2 & 0.1\\
			\midrule
			Unsupervised (Ours) & 29.2 & 72.3 & 0.0 & 0.0 & 0.3 & 0.1 & 4.8 & 59.3 & 38.5 & 77.8 & 25.9 & 42.1\\
			Semi-supervised (Ours) & \textit{\textcolor{blue}{58.9}} & \textit{\textcolor{blue}{78.0}} & \textit{\textcolor{blue}{57.0}} & \textbf{\textcolor{red}{14.1}} & \textbf{\textcolor{red}{53.6}} & \textit{\textcolor{blue}{51.9}} & \textbf{\textcolor{red}{39.1}} & \textit{\textcolor{blue}{79.9}} & \textit{\textcolor{blue}{77.0}} & \textit{\textcolor{blue}{91.0}} & 52.3 & 53.9 \\
			\midrule
			Unsupervised \cite{rochan2022unsupervised} & 34.5 & 54.4 & 15.8 & 3.0 & 1.9 & 27.7 & 7.6 & 65.7 & 57.9 & 82.2 & 29.6 & 34.0 \\
			
			100 Target Frames + \cite{corral2021lidar} & 48.3 & 69.0 & 37.7 & 5.5 & 9.4 & 45.4 & 23.5 & 69.0 & 74.7 & 78.8 & \textit{\textcolor{blue}{56.1}} & \textbf{\textcolor{red}{61.8}} \\
						
			\midrule

			100 Target Frames & 46.3 & 70.3 & 27.1 & 2.0 & 0.1 & 40.3 & 14.7 & 78.1 & 76.0 & 90.7 & 52.1 & 58.0\\
			
			Full Target Dataset$\dagger$  & \textbf{\textcolor{red}{69.5}} &  \textbf{\textcolor{red}{80.0}} &  \textbf{\textcolor{red}{61.7}} & \textit{\textcolor{blue}{11.9}} & \textit{\textcolor{blue}{38.0}} &  \textbf{\textcolor{red}{72.1}} & \textit{\textcolor{blue}{34.2}} &  \textbf{\textcolor{red}{82.6}} &  \textbf{\textcolor{red}{81.4}} &  \textbf{\textcolor{red}{94.0}} &  \textbf{\textcolor{red}{63.7}} &  \textit{\textcolor{blue}{60.7}}\\
			
			
			

			\bottomrule
		\end{tabular}
	}
	\begin{minipage}{0.9\textwidth}\footnotesize
		\vspace{5pt}
		$\dagger$ The target baseline mIoU is lower than reported by the original authors, as we are training from scratch.
	\end{minipage}
	\centering
	\label{tab:k2n_semseg}
\end{table*}
To show how universal our method is, we reversed the domain adaptation of the previous section. We used the training data from the SemanticKITTI dataset to create a fake panoptic segmentation dataset for the lidar sensor of the nuScenes dataset. We trained the Cylinder3D \cite{zhou2020cylinder3d} semantic segmentation network on our fully unsupervised method (only generated frames and pseudo labels) and our semi-supervised approach (all modules of \textsc{Section} \ref{section:method}). We compare both approaches to the fully supervised training on the source and the target dataset in \textsc{Table} \ref{tab:k2n_semseg}.
The semantic segmentation quality of the used network increases with each additional part we outlined in \textsc{Section} \ref{section:method}. 
The naive training on SemanticKITTI data yields a very low performance on the nuScenes validation data of only $7.4$ mIoU. We improve the performance with our unsupervised domain adaptation to a mIoU of $29.2$. We argue, that the lower performance of our unsupervised method on nuScenes compared to the SemanticKITTI dataset is due to the very different vertical aperture angle of the two lidar sensors, i.e. the nuScenes \textit{VLP-32C} lidar sensor has a larger vertical opening angle and can "see" up to $\sim40.73$\,m over the road surface, as opposed to the SemanticKITTI \textit{HDL-64E} sensor which is limited to $\sim3.48$\,m over the ground. Fortunately, most classes of interest are within the mappable area, making the network performance drop less severe than it should be given this large discrepancy.
Our best performing semi-supervised method uses 100 frames of the target dataset, which is $0.36\%$ of the original target training data and reaches  with the use of our pseudo label fusion a final mIoU of $58.9$ as shown in \textsc{Table} \ref{tab:k2n_semseg}. The injection instances are sampled from the same 100 frames to prevent data leakage. Our method therefore enables a performance of $85 \%$ compared to the network trained on the fully labeled target dataset which yields a mIoU of $69.5$. Our semi-supervised method even manages to outperform the fully supervised network on three out of $11$ classes.

We compare both of our approaches with two state of the art domain adaptation methods: our unsupervised approach ($29.2$ mIoU) appears to suffer more from the missing visible areas than the unsupervised domain adaptation of \cite{rochan2022unsupervised} ($34.5$ mIoU). Our semi-supervised approach ($58.9$ mIoU) on the other hand achieves a significant improvement compared to the next best method of \cite{corral2021lidar} with a mIoU of $48.3$ mIoU, as well as their approach with 500 labeled frames ($52.3$ mIoU).

\subsection{NuScenes to Velodyne Alpha Prime}
\label{section:experiments:n2v}
\begin{figure}
	\vspace{4pt}
	\centering
	\includegraphics[width=0.4\textwidth]{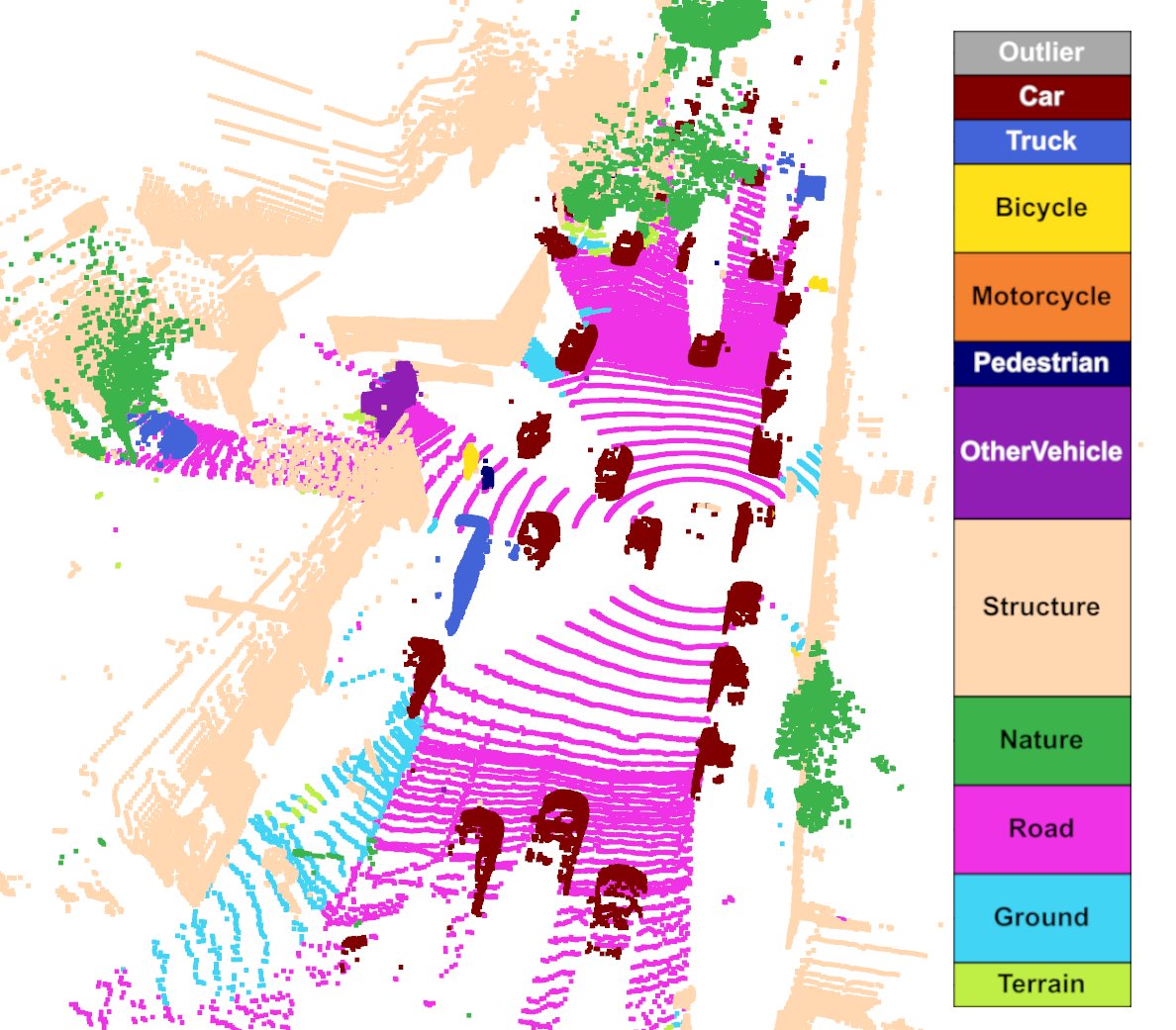}
	\caption{\textbf{Inference Results of the Semantic Segmentation Network Trained on NuScenes Data Recreated in the Structure of the \textit{Velodyne Alpha Prime} Sensor.}}
	\label{fig:velodyne_results}
\end{figure}
We applied our domain adaption method to the training data of the nuScenes dataset in order to recreate a annotated dataset for the high resolution lidar sensor \textit{Velodyne Alpha Prime}. We recorded multiple scenarios in Wuppertal, Germany with this sensor, to produce unlabeled automotive lidar data.
The target lidar has a vertical resolution of 128 non-uniform lidar channels - 4 times the resolution of the lidar sensor used in nuScenes, and a horizontal resolution of 1800 points per scan line, which results in twice the horizontal resolution of the nuScenes lidar data. Also noteworthy is the increased range from $200$\,m to $300$\,m of the target sensor.
We took the same approach as in \textsc{Section} \ref{section:experiments:n2k} and summed up all points of a given scene, to collect as many original lidar measurements as possible. Due to the lower resolution of the source lidar (nuScenes), we ended up with a comparably sparse point cloud. With the meshing process of \textsc{Section} \ref{section:method:mesh} we connected the point cloud to cover the entire visible surrounding. 
Additional to the point cloud generation, we applied two off-the-shelf \cite{mmdet3d2020} 3D bounding box algorithms \cite{lang2019pointpillars} \cite{shi2020points} to unlabeled target data of the \textit{Velodyne Alpha Prime}. As the bounding box detection networks were not trained on this sensor, we filtered out multiple false detections by applying a simple Kalman filter \cite{kalman1960new} to the detections. On these filtered bounding boxes we applied the method of \textsc{Section} \ref{section:method:semi:injections} to sample and inject the lidar points inside of the detected cuboids as semantic instances to our generated training data pool. This two step approach of generating a static point cloud and populating it with real dynamic instances of the target lidar sensor, enabled us to train a well performing network for semantic segmentation of the target lidar data. Qualitative results can be seen in \textsc{Figure} \ref{fig:velodyne_results}. We can not provide a quantitative evaluation, as there is no openly available semantic or panoptic segmentation dataset for the \textit{Velodyne Alpha Prime}.

\subsection{SemanticKITTI to InnovizTwo}
\label{section:experiments:k2i}
\begin{figure}
	\vspace{4pt}
	\centering
	\includegraphics[width=0.4\textwidth]{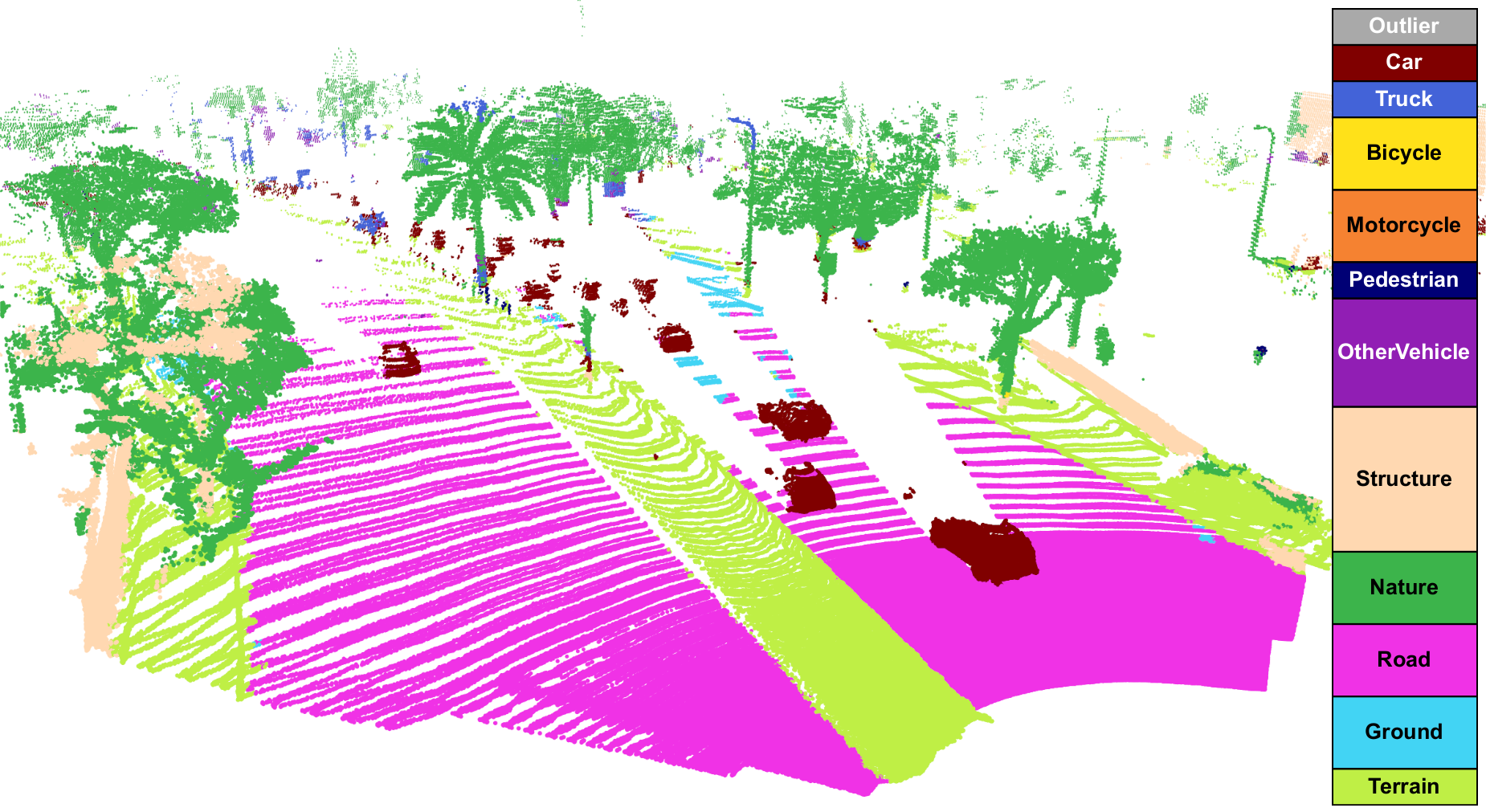}
	\caption{\textbf{Inference Results of the Semantic Segmentation Network Trained on SemanticKITTI Data Recreated in the Structure of the \textit{InnovizTwo} Sensor.}}
	\label{fig:innoviz_results}
\end{figure}
We deliberately chose to apply our domain adaptation to data that do not have segmentation labels to demonstrate the value of our method. The \textit{InnovizTwo} is a high resolution directional lidar sensor with a limited aperture angle in vertical and horizontal direction similar to a depth camera. The range of this sensor is as high as the \textit{Velodyne Alpha Prime} of up to $300$\,m, with a much higher point density in the given direction. Thus, this sensor provides a perfect candidate to perform our domain adaptation from a 360° rotating lidar sensor with a low resolution and range to a directional high resolution lidar to show the generality of our method. The dataset was provided for a self-supervised object detection challenge \cite{innovizeccv}. We use the provided cuboid labels of the 100 annotated frames to define point-wise instances for our semi-supervised domain adaptation as described in \textsc{Section} \ref{section:method:semi:injections}. The results of our trained SalsaNext \cite{cortinhal2020salsanext} semantic segmentation model for the \textit{InnovizTwo} data can be seen in \textsc{Figure} \ref{fig:innoviz_results}.

\section{\uppercase{Conclusion}}

We presented a novel method to recreate annotated segmentation lidar data in the structure of different lidar sensors. 
Furthermore, we have demonstrated the added value of each module of our domain fusion method, to reduce the domain gap between generated and real data, by conducting an extensive ablation study.
Our method works entirely on the data level and can therefore be used with any semantic lidar segmentation model. This is especially useful for future use, as the state of the art for segmentation models is a constantly changing and improving area of research. 
In the future, we plan to extend the application of our method to panoptic segmentation networks as well as 3D bounding box detectors. 
To make our method more robust and to enable the extraction of dynamic objects from source data, we plan to change our mesh creation to require less source point clouds, in order to a) reduce the impact of deviations in the ego-motion and b) to make better use of the source instances for our unsupervised and semi-supervised domain adaptation.

{\small
	\bibliographystyle{apalike}
	\bibliography{Bibliography}
}

%

\end{document}